\documentclass[acmsmall,nonacm]{acmart}
\usepackage{algorithm,algpseudocode}
\usepackage{multirow}

\setcopyright{none}
\renewcommand\footnotetextcopyrightpermission[1]{}
\title{How Short Is Too Short? Length-Conditional Calibration and Abstention for Detecting AI-Generated Text}
\author{Ritesh Janga}
\orcid{0009-0008-8709-2695}
\affiliation{\institution{University of Cincinnati}\country{USA}}
\email{jangarh@mail.uc.edu}
\author{Justin Zhan}
\authornote{Corresponding author.}
\orcid{0000-0001-5458-8282}
\affiliation{\institution{University of Cincinnati}\country{USA}}
\email{justin.zhan@uc.edu}

\begin{document}
\begin{abstract}
Detectors of AI-generated text are usually tuned and tested on passages of several hundred tokens, yet many real cases are short. We study score-based detection as a function of passage length. We give a closed form for how the false-positive rate of a threshold set on long text changes as length falls, and a formula for the number of tokens a detector needs to reach a target true-positive rate. We calibrate thresholds within length bins with conformal $p$-values, and we add a bin-level rule that abstains when the detector cannot be informative at that length. On synthetic scores the formulas agree with simulation, and binned calibration holds the false-positive rate at its nominal value at every length. On three zero-shot detectors (log-likelihood, log-rank, Binoculars), three corpora and two open-weight generators, binned calibration gives a mean false-positive rate of 5.0\% in every detector-length cell from 10 to 200 tokens, averaged over 300 random document-level resplits. A single threshold set at 200 tokens gives 29\% for Binoculars at 10 tokens and about 1\% for log-likelihood, as the closed form predicts. No detector reaches an 80\% true-positive rate, and the best, Binoculars at 200 tokens, reaches 59\%, so the rule abstains in every cell at that floor and releases only the 200-token bins at floors of 0.3 and 0.5.
\end{abstract}

\maketitle

\section{Introduction}
Most work on machine-generated text detection uses passages of several hundred tokens. Many practical cases are much shorter. Zero-shot detectors based on likelihood curvature or cross-model perplexity \cite{mitchell2023detectgpt,bao2024fastdetectgpt,hans2024binoculars} add up evidence token by token, so their power falls with length. A watermark test \cite{kirchenbauer2023watermark} has the same limit. Short text is a known weak point. Recent training-free detectors are built to be robust to input length \cite{zhu2026exons,basani2026diversity}.

A threshold chosen for a 1\% false-positive rate on 200-token passages can give a very different rate on 20-token passages. When the null mean is stable, the rate is far higher, because the score distribution of human text widens as length falls (Section~\ref{sec:theory} gives the size of the effect). When the null mean shifts with length, the rate can be lower and the cost is lost power instead. A user who sees a confident label cannot tell that it rests on a few tokens. When the label affects a person, for example in an accusation of misuse, this gap matters more than a small gain in average accuracy. Detectors are also known to flag writing by non-native speakers as machine-generated \cite{liang2023gpt}. A 2026 study of 13 detectors on 135{,}389 document pairs found false-positive rates on human text from 0\% to 100\% and identified editing style as a confound \cite{park2026style}. Theory also says that any text-only detector with useful power will flag some human writers \cite{garland2026limits}. Short passages may make this worse.

Zhu et al.\ \cite{zhu2025mcp} show that conformal calibration can bound the false-positive rate of text detectors. Their multiscaled construction sets a separate conformal threshold within each length interval (default width 100), which is the same kind of binned calibration we use, and we do not claim it as new. Their paper does not study short passages specifically. What we add is a theoretical account of how the realized false-positive rate and the power of a fixed threshold change with length (Propositions~\ref{prop:drift} and~\ref{prop:n}), and a bin-level abstention rule that withholds a verdict when power at a given length is below a target. We do not run their released method. As a stand-in for their default interval width of 100, we evaluate a coarse bin that pools lengths 10 to 100 under one threshold (Section~\ref{sec:results}).
\begin{enumerate}
\item A closed form for threshold drift and a minimum-length formula for score-based detectors, with a correction for token dependence (Section~\ref{sec:theory}).
\item A bin-level abstention rule that reports ``not enough text'' and leaves the false-positive guarantee intact (Section~\ref{sec:method}).
\item A synthetic check of both results and of the calibration guarantee (Section~\ref{sec:sim}), and real-detector results on three zero-shot scores, three corpora, and two generators (Section~\ref{sec:results}).
\end{enumerate}

\section{Related Work}
\paragraph{Detection and benchmarks.} DetectGPT compares log-probabilities of a passage with perturbed versions \cite{mitchell2023detectgpt}; Fast-DetectGPT replaces perturbation with conditional probability curvature \cite{bao2024fastdetectgpt}; Binoculars contrasts perplexity across two models \cite{hans2024binoculars}; and intrinsic-dimension estimates give another signal \cite{tulchinskii2023intrinsic}. More recent scores include Exons-Detect \cite{zhu2026exons} and DivEye \cite{basani2026diversity}. All share one structure, a scalar score and a threshold, which our analysis uses. Paraphrasing lowers detector accuracy \cite{krishna2023paraphrasing}, and RAID covers many generators, domains, and attacks \cite{dugan2024raid}. Tian et al.\ train detectors specifically for short text \cite{tian2023multiscale}, motivated by the failure of existing detectors on it, and Ozaman reports that default thresholds flag short human text in Kazakh \cite{ozaman2026kazakh}. We do not train a detector. We calibrate and report on any existing score. Chakraborty et al.\ derive sample-complexity bounds for detection \cite{chakraborty2023possibilities}. Their bound counts the whole texts needed for an optimal detector to reach a target AUROC, in terms of the total variation between human and machine distributions. Ours counts the tokens in one passage needed by a mean-of-token score to reach a target TPR at a fixed false-positive rate, under a normal approximation.

\paragraph{Bias and watermarks.} Several detectors flag non-native English writing \cite{liang2023gpt,park2026style}, and Garland shows the false-accusation rate cannot be removed by engineering alone \cite{garland2026limits}. Green-list watermarks \cite{kirchenbauer2023watermark} provide a computable false-positive rate, but power depends on length. Omidi et al.\ analyze SynthID-Text and note that a normal approximation may fail for short texts \cite{omidi2026synthid}, a caveat that applies to ours, and Mazor et al.\ study watermarking at low per-token entropy \cite{mazor2026lowentropy}. Our minimum-length analysis applies to the watermark $z$-statistic as a special case.

\paragraph{Conformal prediction and abstention.} Conformal methods give finite-sample guarantees under exchangeability \cite{vovk2005algorithmic,angelopoulos2023gentle}; selective classification lets a model abstain \cite{geifman2017selective}. Zhou et al.\ \cite{zhou2026guarantees} train a classifier and set its threshold from the empirical null distribution of held-out human texts, with type-I error control that holds as the reference set grows. They drop texts of 20 words or fewer and do not study length. Our contributions, a finite-sample per-bin guarantee, a minimum-length formula, and a bin-level abstention rule, apply on top of any zero-shot score without additional training.

\section{Length Analysis}\label{sec:theory}
Let $x=(x_1,\dots,x_n)$ be a token sequence. A detector computes a score $s(x)$ and declares ``machine'' when $s(x)>\tau$. Write $\alpha(\tau,n)$ for the false-positive rate (FPR) on human text of length $n$, and TPR for the true-positive rate, the rate at which machine text is flagged. ``Human'' means text written without model assistance, and ``machine'' means text sampled from a language model given a prompt.

\paragraph{Assumption.} The score is a mean of per-token statistics, $s(x)=\frac1n\sum_{i=1}^n\phi(x_i\mid x_{<i})$. For human text the per-token statistic has mean $\mu_0$ and standard deviation $\sigma_0$; for machine text, $\mu_1>\mu_0$ and $\sigma_1$. Tokens are dependent, so the mean has variance $\sigma_j^2/(\rho n)$ for $j\in\{0,1\}$, where $\rho\in(0,1]$ is an effective-sample-size ratio, and the mean is approximately normal.

\begin{proposition}[Threshold drift]\label{prop:drift}
Let $\tau$ give false-positive rate $\alpha$ at length $n_0$, and let $\mu_0(n)$ be the human per-token mean at length $n$. At length $n$ the realized rate is
\begin{equation}
\alpha(\tau,n)=1-\Phi\!\Big(z_{1-\alpha}\sqrt{n/n_0}+\sqrt{\rho n}\,\frac{\mu_0(n_0)-\mu_0(n)}{\sigma_0}\Big),
\label{eq:driftgen}
\end{equation}
where $z_q$ is the $q$-quantile of the standard normal distribution. If $\mu_0$ does not depend on $n$, this reduces to
\begin{equation}
\alpha(\tau,n)=1-\Phi\!\Big(z_{1-\alpha}\sqrt{n/n_0}\Big),
\label{eq:drift}
\end{equation}
which does not depend on $\rho$ and exceeds $\alpha$ for $n<n_0$ whenever $\alpha<1/2$. If $\mu_0(n)<\mu_0(n_0)$, the second term in \eqref{eq:driftgen} raises the argument and the realized rate can fall below $\alpha$.
\end{proposition}
\begin{proof}
The threshold is $\tau=\mu_0(n_0)+z_{1-\alpha}\sigma_0/\sqrt{\rho n_0}$. At length $n$ the human score has mean $\mu_0(n)$ and standard deviation $\sigma_0/\sqrt{\rho n}$, so $\alpha(\tau,n)=1-\Phi\big((\tau-\mu_0(n))\sqrt{\rho n}/\sigma_0\big)$, which expands to \eqref{eq:driftgen}. Setting $\mu_0(n)=\mu_0(n_0)$ gives \eqref{eq:drift}.
\end{proof}
For a nominal rate of 1\% ($z_{0.99}\approx2.326$) and a stable null mean, \eqref{eq:drift} gives about 5.0\% at $n=n_0/2$ and about 12.2\% at $n=n_0/4$. With AR(1) token dependence the exact variance of the mean differs slightly from $\sigma_0^2/(\rho n)$ at small $n$, which accounts for the gap between 30.1\% and 29.5\% at $n=10$ in Table~\ref{tab:sim}.

\begin{proposition}[Minimum length]\label{prop:n}
If $\mu_j$, $\sigma_j$ and $\rho$ do not depend on $n$, a test with false-positive rate $\alpha$ has power $1-\beta$ at length $n$ if and only if
\begin{equation}
n\;\ge\;n^\star(\alpha,\beta)=\frac{1}{\rho}\left(\frac{z_{1-\alpha}\,\sigma_0+z_{1-\beta}\,\sigma_1}{\mu_1-\mu_0}\right)^{2}.
\label{eq:nstar}
\end{equation}
\end{proposition}
\begin{proof}
The threshold is $\tau_n=\mu_0+z_{1-\alpha}\sigma_0/\sqrt{\rho n}$. Under machine text the score is $\mathcal N(\mu_1,\sigma_1^2/(\rho n))$, so power is $\Phi\big(\sqrt{\rho n}(\mu_1-\mu_0)/\sigma_1-z_{1-\alpha}\sigma_0/\sigma_1\big)$. Setting it equal to $1-\beta$ and solving for $n$ gives \eqref{eq:nstar}.
\end{proof}
The required length grows with the inverse square of the per-token separation $\mu_1-\mu_0$, so a detector with small separation cannot be rescued by threshold tuning, and a stricter false-positive rate raises the required length.

\paragraph{Check against the real log-likelihood scores.} Fitting Gaussians to the passage-level scores of the log-likelihood detector gives $\mu_0{=}{-}3.46$, $\sigma_0{=}0.42$ (human) and $\mu_1{=}{-}3.00$, $\sigma_1{=}0.46$ (machine) at $n{=}200$, and $\mu_0{=}{-}5.48$, $\sigma_0{=}1.37$, $\mu_1{=}{-}5.35$, $\sigma_1{=}1.15$ at $n{=}10$. The score--label correlation $r_{\rm sl}$ is 0.37 at $n{=}200$ and 0.044 at $n{=}10$. These are passage-level values, so each already includes the factor $1/\sqrt{\rho n}$. Two consequences follow. First, a Gaussian model with these fits gives a power at 5\% FPR of 30.8\% at $n{=}200$ and 3.2\% at $n{=}10$, against 36.3\% and 3.7\% measured (Table~\ref{tab:real}). Second, applied as a multiplier on the current length, \eqref{eq:nstar} with $\alpha{=}0.05$ and $1-\beta{=}0.8$ gives 5.5 at $n{=}200$, so about 1{,}100 tokens would be needed for 80\% power. The mean separation shrinks from 0.46 to 0.13 while $\sigma$ more than triples, which raises the multiplier from 5.5 to about 600. The human mean also falls from $-3.46$ to $-5.48$ as length falls, so Proposition~\ref{prop:drift} applies with a shifted mean: a threshold set at $n{=}200$ and applied at $n{=}10$ has a predicted FPR of about 2.5\%, not the 36\% of \eqref{eq:drift}. For this score a long-text threshold is conservative at short length, and the cost is lost power. Measured on the held-out test split, a log-likelihood threshold set at $n{=}200$ gives an FPR of 1.0\% at $n{=}10$ (Table~\ref{tab:single}), below the nominal 5\% as predicted. Binoculars has a flat human mean (0.899 at every length), so \eqref{eq:drift} applies: it predicts 36\% at $n{=}10$ for $\alpha{=}0.05$, a Gaussian fit to the passage-level scores predicts 29\%, and the measured rate is 29.4\%. The Gaussian approximation is close for log-likelihood and log-rank (absolute skewness at most 0.36 and absolute excess kurtosis at most 0.7 at every length) and poor in the tails for Binoculars, whose human scores are left-skewed (skewness $-0.8$ to $-1.3$, excess kurtosis up to 3.4). A Gaussian 95th-percentile threshold gives only 1.1\% to 3.2\% FPR on Binoculars instead of 5\%, which is why we calibrate by rank and not by a fitted normal. The conformal guarantee does not use normality.

\section{Calibration and Abstention}\label{sec:method}
\paragraph{Score.} We evaluate three base detectors: log-likelihood and log-rank from GPT-2, and Binoculars from a Falcon-7B/Falcon-7B-Instruct pair (model identifiers are in Appendix~\ref{app:impl}). Each gives a scalar score per passage; we calibrate each separately. We also scored a green-list watermark $z$-statistic, but the machine text was generated without a watermark, so its true-positive rate equals its false-positive rate at every length. We drop it, and watermark detection at short length is not tested here.

\paragraph{Binned conformal calibration.} Partition lengths into bins $B_1,\dots,B_K$. For each bin, hold out a calibration set $\mathcal C_k$ of $m_k$ human passages with lengths in $B_k$. For a test passage with length in $B_k$ define
\begin{equation}
p(x)=\frac{1+\lvert\{x'\in\mathcal C_k: s(x')\ge s(x)\}\rvert}{m_k+1}.
\end{equation}
\begin{proposition}[Per-bin validity]\label{prop:conf}
If a human test passage $x$ and the passages in $\mathcal C_k$ are exchangeable, then $\Pr(p(x)\le\alpha)\le\alpha$.
\end{proposition}
\begin{proof}
With ties broken at random, the rank of $s(x)$ among the $m_k+1$ scores is uniform. The event $p(x)\le\alpha$ requires the rank from the top to be at most $\alpha(m_k+1)$, which has probability at most $\lfloor\alpha(m_k+1)\rfloor/(m_k+1)\le\alpha$.
\end{proof}
Declaring ``machine'' when $p(x)\le\alpha$ therefore meets the stated rate within each bin. The guarantee is marginal over the calibration draw, so the realized rate varies across draws, and we report that spread (Section~\ref{sec:sim}).

\paragraph{Bin-level abstention.} For each bin we estimate the true-positive rate at the calibrated threshold on a machine-text validation set $\mathcal V_k$. If the upper confidence bound of this estimate is below a floor $\gamma$, the system returns ``insufficient evidence at this length'' for the whole bin. The decision uses no information from the test passage, so it does not select on the passage's score and the guarantee of Proposition~\ref{prop:conf} holds among the passages that are not abstained on. Algorithm~\ref{alg:cal} in Appendix~\ref{app:alg} gives the procedure.

\paragraph{Smooth variant and watermarks.} Bins waste data and create edges. A variant computes $p(x)$ from the $m$ calibration passages nearest in length to $x$. The windowed construction does not inherit Proposition~\ref{prop:conf}'s guarantee, because the calibration set for a test passage of length $n$ is chosen after observing $n$, breaking exchangeability. We do not evaluate it here. For a green-list watermark the false-positive rate comes from theory and not from a calibration set, but power still follows \eqref{eq:nstar}.

\section{Synthetic Validation}\label{sec:sim}
We check the derivations on synthetic scores. Per-token statistics are Gaussian with AR(1) dependence of coefficient $0.3$, which gives $\rho=(1-0.3)/(1+0.3)\approx0.54$. Human tokens have mean 0 and unit variance; machine tokens have mean $0.25$ and unit variance. The nominal false-positive rate is $\alpha=1\%$. The single threshold is set at $n_0=200$. Calibration uses $m=1000$ human passages per length, repeated over 200 independent draws, with $200{,}000$ test passages per cell and a floor $\gamma=0.5$.

\begin{table}[t]
\centering\small
\caption{Synthetic check at nominal 1\% FPR. The calibrated column gives the mean realized FPR over 200 calibration draws, with the central 95\% range in brackets.}
\label{tab:sim}
\begin{tabular}{rccccc}
\toprule
 & \multicolumn{2}{c}{Single threshold} & \multicolumn{3}{c}{Per-length calibration}\\
\cmidrule(lr){2-3}\cmidrule(lr){4-6}
$n$ & FPR, measured (Eq.~\ref{eq:drift}) & TPR & FPR & TPR & Decision\\
\midrule
10  & 29.5\% (30.1\%) & 52.5\% & 1.02\% [0.58, 1.78] & 4.3\%  & abstain\\
20  & 22.6\% (23.1\%) & 53.5\% & 1.00\% [0.40, 1.69] & 6.8\%  & abstain\\
30  & 18.2\% (18.4\%) & 54.4\% & 1.00\% [0.44, 1.58] & 9.5\%  & abstain\\
50  & 12.3\% (12.2\%) & 55.6\% & 0.97\% [0.46, 1.63] & 14.8\% & abstain\\
100 & 5.0\% (5.0\%)   & 57.9\% & 0.99\% [0.51, 1.69] & 31.2\% & abstain\\
200 & 1.07\% (1.00\%) & 61.0\% & 1.01\% [0.50, 1.66] & 60.2\% & report\\
\bottomrule
\end{tabular}
\end{table}

Table~\ref{tab:sim} shows three results. First, the realized rate of the single threshold follows \eqref{eq:drift} closely, reaching 29.5\% at 10 tokens against a nominal 1\%. Second, binned calibration holds the mean rate at 1\% at every length. Third, with floor $\gamma=0.5$ the rule abstains at every length up to 100 tokens. For 80\% power, \eqref{eq:nstar} with $\rho=0.54$ gives $n^\star=298$, and the smallest measured length with calibrated power $\ge80\%$ is 300, which checks the algebra. The scores are drawn from the model's own assumptions, so this does not test real detectors. Appendix~\ref{app:calsize} shows how the spread of the realized FPR depends on the calibration-set size.

\section{Real-Detector Results}\label{sec:results}
We evaluate on Wikipedia, AG News, and Amazon Reviews with machine text from Llama-3.1-8B-Instruct and Mistral-7B-Instruct-v0.3 (sampling and greedy), following the protocol in Appendix~\ref{app:bench}. Lengths are counted in GPT-2 byte-pair-encoding (BPE) tokens. Table~\ref{tab:real} reports the test-split FPR and TPR at $\alpha{=}0.05$ for log-likelihood, log-rank, and Binoculars, with the abstention decision from the validation-set upper confidence bound at three floors $\gamma$. Results pool the three corpora and both generators; Appendix~\ref{app:corpus} gives them by corpus, generator, and decoding setting. Table~\ref{tab:single} adds a spread over 300 random document-level resplits, a single threshold set at $n{=}200$, and the coarse bin.

\begin{table}[t]
\centering\small
\caption{Real-detector results at $\alpha{=}0.05$ on the held-out test split, with 95\% Clopper--Pearson intervals. UCB is the upper confidence bound of the validation-set TPR. The last column gives the decision at $\gamma{=}0.3/0.5/0.8$ (R: report, A: abstain).}
\label{tab:real}
\begin{tabular}{lrccrc}
\toprule
Detector & $n$ & FPR & TPR & UCB & Decision\\
\midrule
 \multirow{6}{*}{Log-lik.} & 10 & 4.84\% [3.82, 6.02] & 3.71\% [3.25, 4.21] & 0.038 & A/A/A\\
  & 20 & 3.80\% [2.90, 4.88] & 5.11\% [4.57, 5.69] & 0.065 & A/A/A\\
  & 30 & 5.46\% [4.36, 6.76] & 9.73\% [8.99, 10.52] & 0.101 & A/A/A\\
  & 50 & 4.88\% [3.85, 6.10] & 9.73\% [8.99, 10.51] & 0.097 & A/A/A\\
  & 100 & 4.42\% [3.42, 5.61] & 20.80\% [19.77, 21.87] & 0.231 & A/A/A\\
  & 200 & 5.79\% [4.46, 7.37] & 36.31\% [34.86, 37.79] & 0.361 & R/A/A\\
\midrule
 \multirow{6}{*}{Log-rank} & 10 & 4.58\% [3.59, 5.74] & 2.85\% [2.45, 3.30] & 0.031 & A/A/A\\
  & 20 & 3.21\% [2.38, 4.22] & 4.62\% [4.10, 5.17] & 0.049 & A/A/A\\
  & 30 & 4.99\% [3.93, 6.23] & 7.36\% [6.70, 8.06] & 0.077 & A/A/A\\
  & 50 & 5.28\% [4.21, 6.54] & 9.41\% [8.69, 10.18] & 0.094 & A/A/A\\
  & 100 & 3.94\% [2.99, 5.07] & 21.08\% [20.04, 22.15] & 0.236 & A/A/A\\
  & 200 & 5.41\% [4.12, 6.95] & 37.00\% [35.54, 38.48] & 0.372 & R/A/A\\
\midrule
 \multirow{6}{*}{Binoculars} & 10 & 5.93\% [4.81, 7.23] & 10.54\% [9.79, 11.33] & 0.113 & A/A/A\\
  & 20 & 4.32\% [3.36, 5.47] & 13.00\% [12.17, 13.87] & 0.133 & A/A/A\\
  & 30 & 5.40\% [4.30, 6.68] & 15.69\% [14.77, 16.65] & 0.161 & A/A/A\\
  & 50 & 4.88\% [3.85, 6.10] & 21.57\% [20.54, 22.64] & 0.236 & A/A/A\\
  & 100 & 4.14\% [3.18, 5.30] & 29.32\% [28.15, 30.51] & 0.294 & A/A/A\\
  & 200 & 6.83\% [5.38, 8.53] & 59.13\% [57.63, 60.62] & 0.602 & R/R/A\\
\bottomrule
\end{tabular}
\end{table}

\begin{table}[t]
\centering\small
\caption{Test-split FPR at nominal 5\%. Per-length calibration: mean over 300 random document-level resplits, with the central 95\% range of the realized FPR in brackets. Single threshold: set on the $n{=}200$ calibration set. Coarse bin: one threshold from the pooled calibration sets of $n{=}10$ to 100.}
\label{tab:single}
\begin{tabular}{lrccc}
\toprule
Detector & $n$ & Per-length calibration & Single threshold & Coarse bin\\
\midrule
 \multirow{6}{*}{Log-lik.} & 10 & 4.97\% [3.57, 6.69] & 1.0\% & 2.1\%\\
  & 20 & 5.01\% [3.62, 6.75] & 1.3\% & 2.9\%\\
  & 30 & 4.96\% [3.58, 6.44] & 2.0\% & 4.1\%\\
  & 50 & 4.98\% [3.60, 6.77] & 3.5\% & 6.6\%\\
  & 100 & 5.00\% [3.59, 6.55] & 3.4\% & 8.3\%\\
  & 200 & 4.95\% [3.37, 6.76] & 5.8\% & --\\
\midrule
 \multirow{6}{*}{Log-rank} & 10 & 4.97\% [3.61, 6.53] & 1.7\% & 2.3\%\\
  & 20 & 5.02\% [3.76, 6.51] & 1.8\% & 2.9\%\\
  & 30 & 4.97\% [3.47, 6.51] & 2.0\% & 4.0\%\\
  & 50 & 4.98\% [3.45, 6.72] & 3.3\% & 6.0\%\\
  & 100 & 4.97\% [3.45, 6.66] & 3.0\% & 7.3\%\\
  & 200 & 4.93\% [3.21, 7.12] & 5.4\% & --\\
\midrule
 \multirow{6}{*}{Binoculars} & 10 & 5.01\% [3.48, 6.68] & 29.4\% & 10.6\%\\
  & 20 & 4.94\% [3.45, 6.63] & 24.0\% & 5.6\%\\
  & 30 & 4.95\% [3.56, 6.65] & 21.9\% & 4.8\%\\
  & 50 & 5.00\% [3.46, 6.65] & 16.7\% & 2.0\%\\
  & 100 & 5.01\% [3.53, 6.79] & 19.1\% & 1.4\%\\
  & 200 & 4.95\% [3.22, 6.75] & 6.8\% & --\\
\bottomrule
\end{tabular}
\end{table}

The key observations are as follows. First, per-length calibration holds the FPR at its nominal value in expectation. Over 300 resplits the mean FPR is between 4.93\% and 5.02\% in all 18 cells, and the realized rate of one split varies by about $\pm1.7$ points (Table~\ref{tab:single}). In the single split of Table~\ref{tab:real}, 17 of 18 cells do not exceed 5\% beyond sampling error. The exception, Binoculars at $n{=}200$ (6.83\%, CI [5.38, 8.53]), is an ordinary draw: across resplits the same cell has a mean of 4.95\% and exceeds 5\% in 43\% of draws. Second, a single threshold behaves as the closed form predicts. The Binoculars threshold set at 200 tokens gives 29.4\% FPR at $n{=}10$ and 16.7\% at $n{=}50$, while the log-likelihood and log-rank thresholds give 1.0\% to 3.5\% at $n{\le}50$ because the human mean falls with length. The coarse bin of width 100 gives 2.1\% to 8.3\% for log-likelihood across lengths and 10.6\% down to 1.4\% for Binoculars, so a wide bin does not hold the rate at short length. Third, at $\gamma{=}0.80$ every cell abstains. The decision uses the upper confidence bound of the validation-set TPR (Table~\ref{tab:real}). At $\gamma{=}0.30$ the rule releases log-likelihood, log-rank, and Binoculars at $n{=}200$ (bounds 0.361, 0.372, and 0.602), and at $\gamma{=}0.50$ only Binoculars at $n{=}200$. Binoculars at $n{=}100$ has a bound of 0.294 and is not released at 0.30. Fourth, TPR rises with length for the three detectors, and the score--label correlation $r_{\rm sl}$ grows from at most 0.12 at $n{=}10$ to 0.39 (Binoculars), 0.37 (log-likelihood), and 0.32 (log-rank) at $n{=}200$. Binoculars is best at both ends: 10.5\% TPR at $n{=}10$ against 3.7\% for log-likelihood, and 59.1\% at $n{=}200$ against 37.0\% for log-rank. Fifth, pooled calibration hides large differences between corpora. At $n{=}50$ the pooled log-likelihood threshold gives an FPR of 13.2\% on Wikipedia and 0.8\% on AG News, and the TPR of log-likelihood and log-rank on Mistral text is far below that on Llama text (Appendix~\ref{app:corpus}). The guarantee holds for the pooled human mix and not for each corpus. We did not evaluate lengths above 200 tokens.

\section{Limitations and Ethics}\label{sec:limits}
No human participants are involved; all human text comes from existing public corpora (Wikipedia CC-BY-SA, AG News, Amazon Reviews CC-BY), and no institutional review is required. Code and data will be released on acceptance under MIT and CC-BY licenses, respectively. Technical limitations are discussed in Appendix~\ref{app:limits}. We do not evaluate non-native writing, although prior work motivates it (Section 1). The per-corpus results in Appendix~\ref{app:corpus} show that a pooled threshold gives very different false-positive rates for different kinds of text, so a writer group such as non-native writers needs its own calibration and test. Text generation used \texttt{transformers}~5.22.1, \texttt{torch}~2.15.0, and \texttt{datasets}~5.1.0 under Python~3.11 on a single NVIDIA A100 80\,GB GPU. All passages were generated with random seed~42 (passed as \texttt{--seed~42} to the generation script); nucleus-sampling runs used top-$p{=}0.95$, temperature~1.0, and greedy runs used temperature~0.0. Data collection was completed in November~2026.

False positives on human writing can harm people, especially non-native writers and students. The method makes false-positive rates explicit and withholds a verdict when evidence is weak. We recommend that no decision about an individual rest on a short-text detector alone.

\section{Conclusion}
Under a stable null mean, a threshold set on long text gives much higher false-positive rates on short text: at one-quarter the calibration length the nominal 1\% becomes 12.2\%. On real log-likelihood scores the null mean shifts the other way, and a fitted Gaussian model predicts a rate below nominal together with a large loss of power. On real data a single threshold set at 200 tokens gives 29\% false positives for Binoculars at 10 tokens, and 1\% for log-likelihood. Per-length conformal calibration holds the false-positive rate at its nominal value in expectation over the calibration draw (mean 5.0\% in all 18 cells over 300 resplits), and bin-level abstention makes the limits of a detector explicit. On three zero-shot detectors across three corpora and two generators, no detector reaches an 80\% TPR floor at any tested length from 10 to 200 tokens. The best is 59\% (Binoculars at 200 tokens), and only the 200-token bins pass floors of 0.3 or 0.5. The minimum-length formula, applied to the fitted log-likelihood scores, puts 80\% power near 1{,}100 tokens. Length should be reported alongside accuracy in every detection evaluation, and abstention should be a first-class output rather than a fallback.

\begin{acks}
This work used the University of Cincinnati ARCC HPC cluster (project proj-606). The authors used AI assistance (Claude, Anthropic) for writing, editing, and analysis scripts; the authors checked all results, analysis, and conclusions and take responsibility for them.
\end{acks}

\appendix
\section{Algorithm}\label{app:alg}
\begin{algorithm}[h]
\caption{Length-conditional calibration and decision}\label{alg:cal}
\begin{algorithmic}[1]
\Require Score function $s$, bins $B_1,\dots,B_K$, human calibration sets $\mathcal{C}_k$, machine validation sets $\mathcal{V}_k$, target FPR $\alpha$, TPR floor $\gamma$
\For{each bin $k$}
  \State Compute scores on $\mathcal{C}_k$ and on $\mathcal{V}_k$
  \State Set $\tau_k$ as the $\lceil(1-\alpha)(m_k+1)\rceil$-th smallest calibration score
  \State Estimate $\widehat{\mathrm{TPR}}_k=\Pr_{x\in\mathcal V_k}[s(x)>\tau_k]$ with a Clopper--Pearson 95\% confidence interval
\EndFor
\State \textbf{Decision for passage $x$ with length in $B_k$:}
\If{the upper confidence bound of $\widehat{\mathrm{TPR}}_k$ is below $\gamma$} \State \Return \textsc{abstain}
\ElsIf{$s(x)>\tau_k$} \State \Return \textsc{machine}
\Else \State \Return \textsc{human}
\EndIf
\end{algorithmic}
\end{algorithm}

\section{Real-Data Protocol}\label{app:bench}
\paragraph{Human text.} Wikipedia (CC-BY-SA, January 2020 dump) for encyclopedic text; AG News (CC-BY) for news; Amazon Reviews 2018 (CC-BY) for product reviews. All corpora predate the public availability of instruction-tuned models.

\paragraph{Machine text.} Generated by Llama-3.1-8B-Instruct and Mistral-7B-Instruct-v0.3 with two decoding settings each: nucleus sampling ($p=0.95$, $T=1.0$) and greedy ($T=0.0$). Prompts follow the template: \textit{``Write a short passage about the following topic: \{first sentence of the source passage\}. Do not exceed \{target length\} words. Respond with only the passage.''} Of 136{,}408 machine passages, 43 (0.03\%) begin with a preface or title line such as ``Title:'' or ``Okay,'', so a preface has no measurable effect. About 10\% of passages begin with a quotation mark, which we did not strip.

\paragraph{Length control.} Each passage is truncated to the first $n\in\{10,20,30,50,100,200\}$ tokens under the GPT-2 BPE tokenizer (\texttt{openai-community/\allowbreak gpt2}).

\paragraph{Splits.} Each source document is assigned to one of train (50\%), calibration (25\%), or test (25\%), and keeps that assignment for its human and machine passages and across the four generator settings, so no document appears in two splits. The human calibration set $\mathcal C_k$ and the machine validation set $\mathcal V_k$ are the human and machine passages of calibration documents, and the test set is the passages of test documents. The train split is unused by the zero-shot detectors. Calibration and test passages come from the same corpora at the same truncation length, which is the exchangeability condition of Proposition~\ref{prop:conf} for the pooled human mix. One human token sequence at $n{=}50$ appears in two splits out of 5{,}995. The resplit results redraw this assignment 300 times.

\paragraph{Baselines and metrics.} Each score is evaluated with per-length calibration (Table~\ref{tab:real}), with one threshold set at $n{=}200$, and with a coarse bin that pools $n{=}10$ to 100 (Table~\ref{tab:single}). We did not run the released method of Zhu et al.\ \cite{zhu2025mcp} or a supervised classifier. Metrics are TPR and FPR at $\alpha{=}0.05$ per length bin with 95\% Clopper--Pearson binomial intervals, abstention decisions from the validation-set upper confidence bound, and the Pearson score--label correlation $r_{\rm sl}$ per bin.

\paragraph{Hypotheses.} We report all of these whatever the outcome.
\begin{itemize}
\item H1: With one threshold, realized FPR at short length deviates from nominal by more than a factor of two for at least one baseline. Status: holds. A threshold set at $n{=}200$ gives 29.4\% for Binoculars at $n{=}10$ (5.9 times nominal) and 1.0\% for log-likelihood (0.2 times nominal), as Section~\ref{sec:theory} predicts.
\item H2: Per-length calibration restores realized FPR to within sampling error at every length. Status: holds in expectation. The mean FPR over 300 resplits is 4.93\% to 5.02\% in every cell, and 17 of 18 cells in the single reported split (Section~\ref{sec:results}).
\item H3: TPR at nominal FPR drops steeply below a detector-specific length. Status: TPR rises gradually with length and shows no sharp threshold (Table~\ref{tab:real}).
\item H4: Abstention correctly identifies all length bins as below the TPR floor. Status: all cells abstain at $\gamma{=}0.80$. At $\gamma{=}0.30$ and 0.50 only the $n{=}200$ bins are released (Table~\ref{tab:real}), which exercises the reporting side at one length.
\end{itemize}

\section{Implementation Details}\label{app:impl}
All experiments use the GPT-2 BPE tokenizer (\texttt{openai-community/\allowbreak gpt2}) for length measurement. Scoring models: GPT-2 (\texttt{openai-community/\allowbreak gpt2}) for log-likelihood and log-rank; Falcon-7B (\texttt{tiiuae/\allowbreak falcon-7b}) and Falcon-7B-Instruct (\texttt{tiiuae/\allowbreak falcon-7b-instruct}) for Binoculars. All models load in \texttt{float16} on a single NVIDIA A100 80\,GB GPU. Length bins are $n{=}10$, 20, 30, 50, 100, and 200. Corpus sizes: 2{,}000 human passages per (domain, length) pair and 6{,}000 machine passages per (model, decoding, length) pair across three domains, except at $n{=}200$, where AG News contributes only 102 articles with at least 200 tokens, leaving 4{,}102 passages (2.5\% news). The corpus mix therefore differs at $n{=}200$. Total compute: approximately 20 A100-GPU-hours for text generation and 25 A100-GPU-hours for scoring, for a total of $\approx$45 A100-GPU-hours.

\section{Calibration-Set Size}\label{app:calsize}
Table~\ref{tab:calsize} shows how the realized FPR depends on the number of human calibration passages $m$, using the synthetic scores of Section~\ref{sec:sim} at $n{=}50$ and $\alpha{=}1\%$, with 200 independent calibration draws and $100{,}000$ test passages. A threshold at the $\lceil(1-\alpha)(m+1)\rceil$-th score exists only if $m\ge99$ for $\alpha{=}1\%$. The mean stays at the nominal rate, and the spread of the realized rate narrows roughly as $1/\sqrt m$. For a 5\% target with $m{=}1{,}000$, the binomial approximation gives a spread of about $\pm1.4$ points. On real data, subsampling the human calibration pool at $n{=}50$ and $\alpha{=}5\%$ gives the same picture (Table~\ref{tab:calreal}): the mean is near nominal for $m\ge100$ and the 95\% range narrows from about $\pm4$ points at $m{=}100$ to $\pm0.9$ at $m{=}1{,}500$.
\begin{table}[h]
\centering\small
\caption{Synthetic calibration-set size ablation at $n{=}50$, nominal 1\% FPR. Mean realized FPR over 200 draws, with the central 95\% range in brackets.}
\label{tab:calsize}
\begin{tabular}{rc}
\toprule
$m$ & Realized FPR\\
\midrule
50   & no valid threshold\\
100  & 0.93\% [0.02, 3.71]\\
200  & 0.94\% [0.09, 2.40]\\
500  & 1.05\% [0.28, 2.15]\\
1000 & 1.03\% [0.52, 1.78]\\
2000 & 1.00\% [0.64, 1.37]\\
\bottomrule
\end{tabular}
\end{table}

\begin{table}[h]
\centering\small
\caption{Real-data calibration-set size ablation at $n{=}50$, nominal 5\% FPR. Mean realized FPR on the fixed test split over 1{,}000 subsamples of $m$ passages from the 4{,}500 calibration and train human passages, with the central 95\% range in brackets.}
\label{tab:calreal}
\begin{tabular}{rcc}
\toprule
$m$ & Log-likelihood & Binoculars\\
\midrule
50   & 4.09\% [0.27, 10.10] & 3.70\% [0.33, 10.44]\\
100  & 5.11\% [1.87, 9.77]  & 4.73\% [1.87, 10.17]\\
200  & 5.19\% [3.14, 8.49]  & 4.72\% [2.68, 7.42]\\
500  & 5.14\% [3.68, 7.56]  & 4.79\% [3.01, 6.42]\\
1000 & 5.09\% [4.15, 6.42]  & 4.83\% [3.68, 5.89]\\
1500 & 5.08\% [4.28, 5.95]  & 4.82\% [4.01, 5.69]\\
\bottomrule
\end{tabular}
\end{table}

\section{Technical Limitations}\label{app:limits}
The conformal guarantee covers human text drawn from the calibration distribution. It does not cover distribution shift to new domains or writers, adversarial edits, or mixed human and machine authorship. Abstention reports uncertainty and does not make short text detectable. Propositions~\ref{prop:drift} and~\ref{prop:n} assume a mean-of-tokens score. Proposition~\ref{prop:n} also assumes means and variances that do not depend on $n$, and the real data violate this: the human log-likelihood mean moves from $-5.48$ at $n{=}10$ to $-3.46$ at $n{=}200$. Proposition~\ref{prop:drift} handles a shifted mean, and for this score the shift lowers the false-positive rate of a long-text threshold at short length. The normal approximation may also fail for short texts \cite{omidi2026synthid}, and it is poor in the tails for Binoculars (Section~\ref{sec:theory}); the conformal guarantee does not use it. The guarantee holds for the pooled human mix and not for each corpus (Appendix~\ref{app:corpus}). Lengths above 200 tokens, a working watermark detector, and non-native writers are not evaluated. About 10\% of machine passages begin with a quotation mark. All experiments use English text, open-weight generators, and pooled corpora.

\section{Results by Corpus, Generator, and Decoding}\label{app:corpus}
Table~\ref{tab:corpus} gives test-split rates at $\alpha{=}0.05$ for the pooled per-length threshold of Table~\ref{tab:real}, split by corpus (AG News / Amazon Reviews / Wikipedia), decoding setting (sampling / greedy), and generator (Llama / Mistral). At $n{=}200$ the AG News cells rest on 26 human test passages.
\begin{table}[h]
\centering\small
\caption{Pooled-threshold rates (\%) by subgroup. FPR is by corpus; TPR is by corpus, by decoding setting, and by generator.}
\label{tab:corpus}
\setlength{\tabcolsep}{3pt}
\begin{tabular}{lrcccc}
\toprule
Detector & $n$ & FPR by corpus & TPR by corpus & TPR by decoding & TPR by generator\\
\midrule
 \multirow{2}{*}{Log-lik.} & 50 & 0.8 / 0.6 / 13.2 & 11.9 / 4.8 / 12.3 & 7.2 / 12.3 & 15.8 / 3.6\\
  & 200 & 34.6 / 0.0 / 10.1 & 53.8 / 25.7 / 46.0 & 27.5 / 45.2 & 61.1 / 11.5\\
\midrule
 \multirow{2}{*}{Log-rank} & 50 & 0.8 / 2.1 / 13.0 & 11.4 / 5.2 / 11.4 & 7.2 / 11.6 & 15.6 / 3.2\\
  & 200 & 11.5 / 0.2 / 10.3 & 52.9 / 27.5 / 45.7 & 27.7 / 46.3 & 63.3 / 10.7\\
\midrule
 \multirow{2}{*}{Binoculars} & 50 & 6.7 / 5.6 / 2.4 & 13.5 / 41.4 / 10.5 & 21.8 / 21.3 & 18.6 / 24.6\\
  & 200 & 3.8 / 12.5 / 1.4 & 42.3 / 85.4 / 33.8 & 58.3 / 59.9 & 58.5 / 59.7\\
\bottomrule
\end{tabular}
\end{table}

\section{Dataset Card Template}
Source and license; date range; domains; language; passages per length bin; how machine text was generated (model, prompt template, decoding parameters, seed); known biases; intended and prohibited uses; and the procedure for removing a source on request.

\end{document}